\documentclass{anstrans}
\title{Development and Deployment of Hybrid ML Models for Critical Heat Flux Prediction in Annulus Geometries}
\author{Aidan Furlong,$^{*}$ Xingang Zhao,$^{\dagger}$ Robert Salko,$^{\ddagger}$ Xu Wu$^{*}$}

\institute{
$^{*}$Department of Nuclear Engineering, North Carolina State University, Burlington Engineering Laboratories, 2500 Stinson Drive, Raleigh, NC 27695, ajfurlon@ncsu.edu, xwu27@ncsu.edu
\and
$^{\dagger}$Department of Nuclear Engineering, University of Tennessee, Knoxville, Zeanah Engineering Complex, 863 Neyland Dr, Knoxville, TN 37996, xzhao47@utk.edu
\and
$^{\ddagger}$Nuclear Energy and Fuel Cycle Division, Oak Ridge National Laboratory Oak Ridge, TN, USA 37830, salkork@ornl.gov
}

\usepackage{graphicx} 
\usepackage{booktabs} 
\usepackage{microtype} 
\usepackage{siunitx}

\newcommand\blfootnote[1]{%
  \begingroup
  \begin{NoHyper}
  \renewcommand\thefootnote{}\footnote{#1}%
  \addtocounter{footnote}{-1}%
  \end{NoHyper}
  \endgroup
}

\begin{document}
\section{Introduction}
\blfootnote{This manuscript has been authored in part 
by UT-Battelle, LLC, under contract DE-AC05-00OR22725 
with the US Department of Energy (DOE). The publisher 
acknowledges the US government license to provide public 
access under the DOE Public Access Plan 
(https://energy.gov/doe-public-access-plan).}

Accurate prediction of critical heat flux (CHF) is an essential component of safety analysis in pressurized and boiling water reactors. When a nuclear system exceeds the CHF value, a temperature excursion in fuel rod cladding occurs. If large enough in magnitude, such an event has the potential to cause cladding failure and fuel melt. Preventing this condition requires a precise calculation of CHF during thermal hydraulic analysis. To support this goal, several empirical correlations and lookup tables have been constructed from physical experiments over
the past several decades. These tools are typically based loosely on a combination of fundamental physics and curve fits, and they can show significant deviation from experimental data in various operational regions.

With the onset of accessible machine learning (ML) frameworks,  several initiatives have been established with the goal of predicting CHF more accurately than the aforementioned, older tools. While purely data-driven surrogate modeling has been extensively investigated, these approaches lack interpretability, lack resilience to data scarcity, and have been developed mostly using data from tube experiments. Therefore, a bias-correction hybrid approach has been developed~\cite{zhao2020prediction}. Hybrid models rely on an initial ``low-fidelity'' estimate provided by a base model such as an empirical correlation, which is then corrected by a data-driven ML model trained on experimental–base model residuals. Several studies have noted improved accuracy and behavior in various operational conditions when compared to the base models and purely data-driven surrogates, all considering tube geometries. Our previous work~\cite{furlong2025deployment} successfully deployed hybrid tube-based models in the CTF subchannel code~\cite{salko2020ctf}, supporting the ongoing initiative to improve CTF's performance in dryout experiments.

This study focuses on developing and deploying \textit{annulus-specific CHF capabilities} for use in CTF, which currently do not exist in the code. It is well known that CHF behavior can differ greatly in annuli compared to tubes; instead of attempting to create one unified model capable of predicting both tubes and annuli, a set of specialized annuli-specific ML models were constructed. Three base models are used to accomplish this: the Biasi~\cite{biasi1967studies}, Bowring~\cite{bowring1972simple}, and Katto~\cite{katto1979generalized} empirical correlations. On top of these, three corresponding hybrid ML models were built in addition to a pure ML model for comparison. This study describes the construction of these annulus models, their implementation within CTF, and the validation performed using the modified CTF code.

\section{Methods}

\subsection{Empirical Correlations}

The Biasi and Bowring correlations were chosen because they are commonly used in subchannel codes, and they are currently implemented in CTF using the direct substitution method. For this study, since the annuli are assumed to be isolated subchannels, the heat balance method was used~\cite{hejzlar1996consideration}. The chosen Katto correlation is a modified version of the tube-derived model for use in annuli. It has been argued~\cite{katto1979generalized} that the heated equivalent diameter, $D_{\mathrm{he}}$, is more appropriate than the hydraulic diameter, $D_{\mathrm{hy}}$, when extending tube-derived correlations to annuli. This is physically motivated, since the channel power equals the axial integral of heat flux over the \textit{heated} perimeter and flow length. Therefore, it is the heated rather than the total wetted perimeter that should define the characteristic diameter, resulting in the use of $D_{\mathrm{he}}$.

For an annulus with a heated inner wall, the heated equivalent diameter is computed with Equation (\ref{eqn:heated_equivalent_diameter}). Here, $A_{\mathrm{sc}}$ is the flow area of the subchannel, $P_{\mathrm{he}}$ is the heated perimeter, and $d_{\mathrm{o}}$ and $d_{\mathrm{i}}$ are the outer and inner wall diameters, respectively. Note that $D_{\mathrm{he}}$ converges to $D_{\mathrm{hy}}$ as the heated perimeter approaches the total wetted perimeter.

\begin{equation} \label{eqn:heated_equivalent_diameter}
    D_{\mathrm{he}} = \frac{4A_{\mathrm{sc}}}{P_{\mathrm{he}}} = \frac{d^2_{\mathrm{o}}-d^2_{\mathrm{i}}}{d_{\mathrm{i}}}
\end{equation}

\subsection{Hybrid ML Strategy}

\begin{figure}[ht!]
    \centering
    \includegraphics[width=0.6\linewidth]{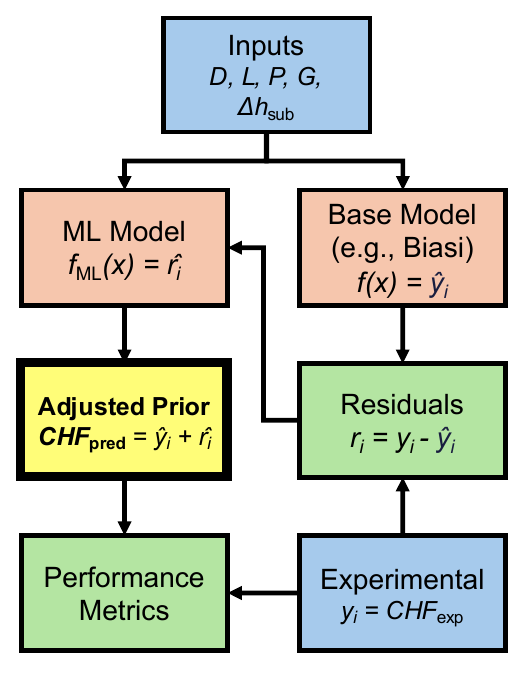}
    \caption{Hybrid ML bias-correction strategy in \textit{training} configuration.}
    \label{fig:hybrid_workflow}
\end{figure}

All ML-based models used five input features: heated equivalent diameter ($D_{\mathrm{he}}$), heated length ($L$), pressure ($P$), mass flux ($G$), and inlet subcooling ($\Delta h_{\mathrm{sub,in}}$). In the hybrid model framework, a low-fidelity model provides an initial estimate, which is then corrected by an ML-predicted residual. This configuration embeds physics-informed structure into the prediction pipeline, reducing the ML model’s burden and improving interpretability. The \textit{training} workflow is depicted in Figure \ref{fig:hybrid_workflow}, when experimental data is available. The base model's output ($\hat{y}_i$) is compared against the experimental data ($y_i$) to compute a residual ($\hat{r_i}$), which is then used to train the ML model. The final CHF prediction is formed by adding the predicted residual ($\hat{r}_i$) to the base model estimate.

The architecture and the hyperparameters of the models themselves were retained from previous studies that used data from tube geometries, given the similarity of the tasks. Re-optimizing the model architecture would require a significant amount of data to be allocated in an isolated validation dataset, only to be used during tuning to prevent information leakage. The transferred model structure consisted of seven fully-connected hidden layers, with all weight/bias initializations re-seeded, as this study is not focused on a transfer learning. 

To reduce overfitting, a learning rate decay and early stopping strategy were implemented. Learning rate decay was configured to exponentially decay with a factor of 0.96 every epoch, which promotes finer tuning as training progresses. Early stopping uses a small validation partition taken from the training data to terminate training once the validation loss fails to improve over a set amount of time---in this case, 25 epochs. Each of the models was permitted to train to a maximum of 500 epochs. After training, the training/validation loss curves were inspected for signs of overfitting. Because all models were trained in TensorFlow, they were exported in HDF5 format and loaded into CTF using a custom Fortran-native framework~\cite{furlong2025native}.

\subsection{Data Processing}

Four datasets were compiled for this study, all from annulus geometry CHF experiments. They cover a range of operational conditions, which are provided in Table \ref{tab:dataset_info}. It should be noted that Beus, Janssen, and Mortimore consider uniformly internally heated, vertical upflow arrangements using water as the working fluid. In the Becker dataset, 23 of its points are uniformly heated, and the remaining points use various heating profiles. The Becker experiments found that the use of differing power profiles had little effect on the experimental CHF values. Therefore, we included all the Becker data points in order to have more training data.

\begin{table*}[ht!]
    \centering
    \caption{Ranges of experimental conditions for the four compiled annulus datasets. The outlet equilibrium quality is denoted by $x_{\mathrm{e,cr}}$, and $q''_{\mathrm{cr}}$ represents the critical heat flux. Values for $G$ and $q''_{\mathrm{cr}}$ are provided with the precision reported in the original text. Note that the upper value in a given parameter cell indicates the \textbf{maximum}, and the lower value indicates the \textbf{minimum}.}
    \begin{tabular}{lcccccccc}
    \toprule
        Dataset & Points & $D_{\mathrm{he}}$ & $L$ & $P$ & $G$ & $\Delta h_{\mathrm{sub,in}}$ & $x_{\mathrm{e,cr}}$ & $q''_{\mathrm{cr}}$ \\
                  & & (\si{\milli\meter}) & (\si{\meter}) & (\si{\mega\pascal}) & (\si{\kilo\gram\per\square\meter\per\second}) & (\si{\kilo\joule\per\kilo\gram}) & (-) & (\si{\kilo\watt\per\square\meter})\\ \midrule
        Becker \cite{becker1981experimental}   & 199 & 21.82 & 3.60 & 7.04  & 2496 & 206.84  &  0.57 & 2025 \\
                                               &     & 21.82 & 2.95 & 6.56  & 249  & 45.82   &  0.16 & 323  \\ \midrule
        Beus \cite{beus1981critical}           & 77  & 15.20 & 2.13 & 15.55 & 3721 & 1163.03 &  0.23 & 3300 \\
                                               &     & 15.20 & 2.13 & 5.52  & 671  & 135.33  & -0.31 & 800  \\ \midrule       
        Janssen \cite{janssen1963burnout}      & 282 & 96.30 & 2.74 & 9.72  & 5913 & 950.10  &  0.21 & 6000 \\
                                               &     & 11.30 & 0.74 & 4.13  & 381  & 6.98    & -0.13 & 1400 \\ \midrule
        Mortimore \cite{mortimore1979critical} & 19  & 13.30 & 2.13 & 13.79 & 3637 & 1159.27 &  0.22 & 2300 \\
                                               &     & 13.30 & 2.13 & 8.27  & 677  & 131.09  & -0.13 & 900  \\ \midrule
        \textbf{All Data}                      & 577 & 96.30 & 3.60 & 15.55 & 5913 & 1163.03 &  0.22 & 6000 \\
                                               &     & 11.30 & 0.74 & 4.13  & 249  & 6.98    & -0.13 & 323  \\
    \bottomrule
    \end{tabular}
    \label{tab:dataset_info}
\end{table*}

\begin{figure}[ht!]
    \centering
    \includegraphics[width=\linewidth]{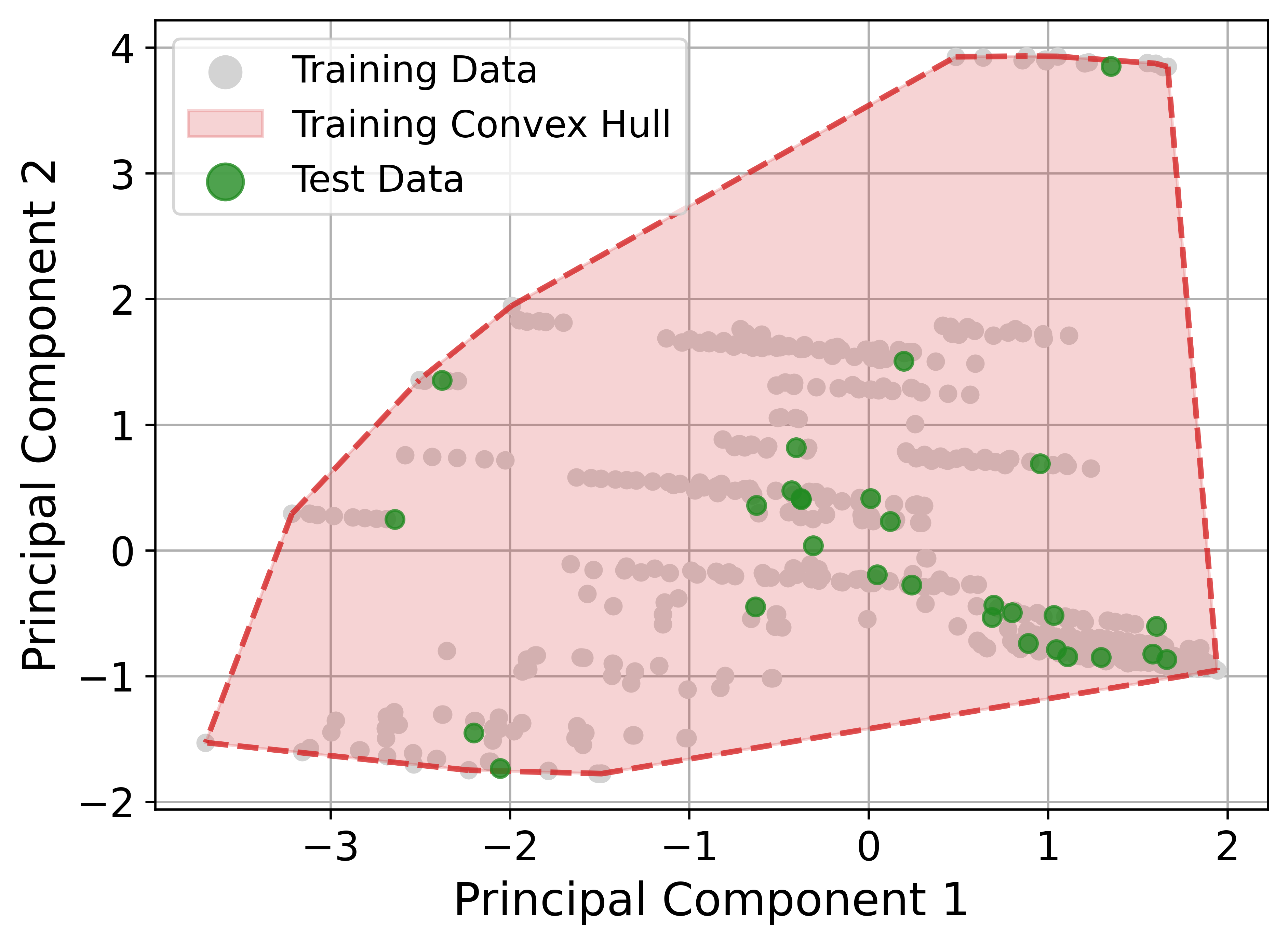}
    \caption{Visualization of the test data points distribution over the training domain. All test points are in an interpolating regime, as indicated by their location on the convex hull defined by the PCA transformation.}
    \label{fig:general_data_pca}
\end{figure}

During data extraction, all entries were used to compute the base model CHF estimates and, subsequently, the experimental--estimate residuals for hybrid model training. The combined dataset was then standardized, shuffled, and partitioned using a 90/5/5 split for training, validation, and testing, respectively. Given the limited data, maximizing the training fraction was necessary, so  a small test set (29 points) was used. To ensure that the test set provided a representative assessment of model generalization, its coverage of the training domain was examined. Poor sampling could lead to misleading performance metrics, if the test points occupy a small subregion. To visualize the distribution of testing instances, a two-component principal component analysis (PCA) was performed; the results are plotted in Figure \ref{fig:general_data_pca}. A convex hull was also computed to verify that all test points fall within the training distribution, as evaluation of out-of-distribution behavior is outside the scope of this study.

\section{Results}

To evaluate the performance of empirical correlations and ML models within CTF, uniformly heated subchannel models were constructed. Each simulation consisted of 60 axial nodes and was run as a transient to steady state over a 40-second simulation window. Prior convergence studies have confirmed that this spatial and temporal resolution is sufficient to achieve stable exit node conditions. CHF was computed at the exit node by multiplying the local departure from nucleate boiling ratio by the pin surface heat flux.

The standalone empirical correlations were first applied to the test partition to establish a performance baseline. Five performance metrics were computed for model/data comparison and are reported in Table \ref{tab:comparison_base}. The metrics of interest were the mean absolute relative error ($\mu_\text{error}$), maximum relative error ($\text{Max}_{\text{error}}$), standard deviation of the relative error ($\text{Std}_{\text{error}}$), relative root-mean-square-error ($rRMSE$), and the fraction of points with error values exceeding 10\% ($F_{\text{error}}>10\%$).

\begin{figure}
    \centering
    \includegraphics[width=\linewidth]{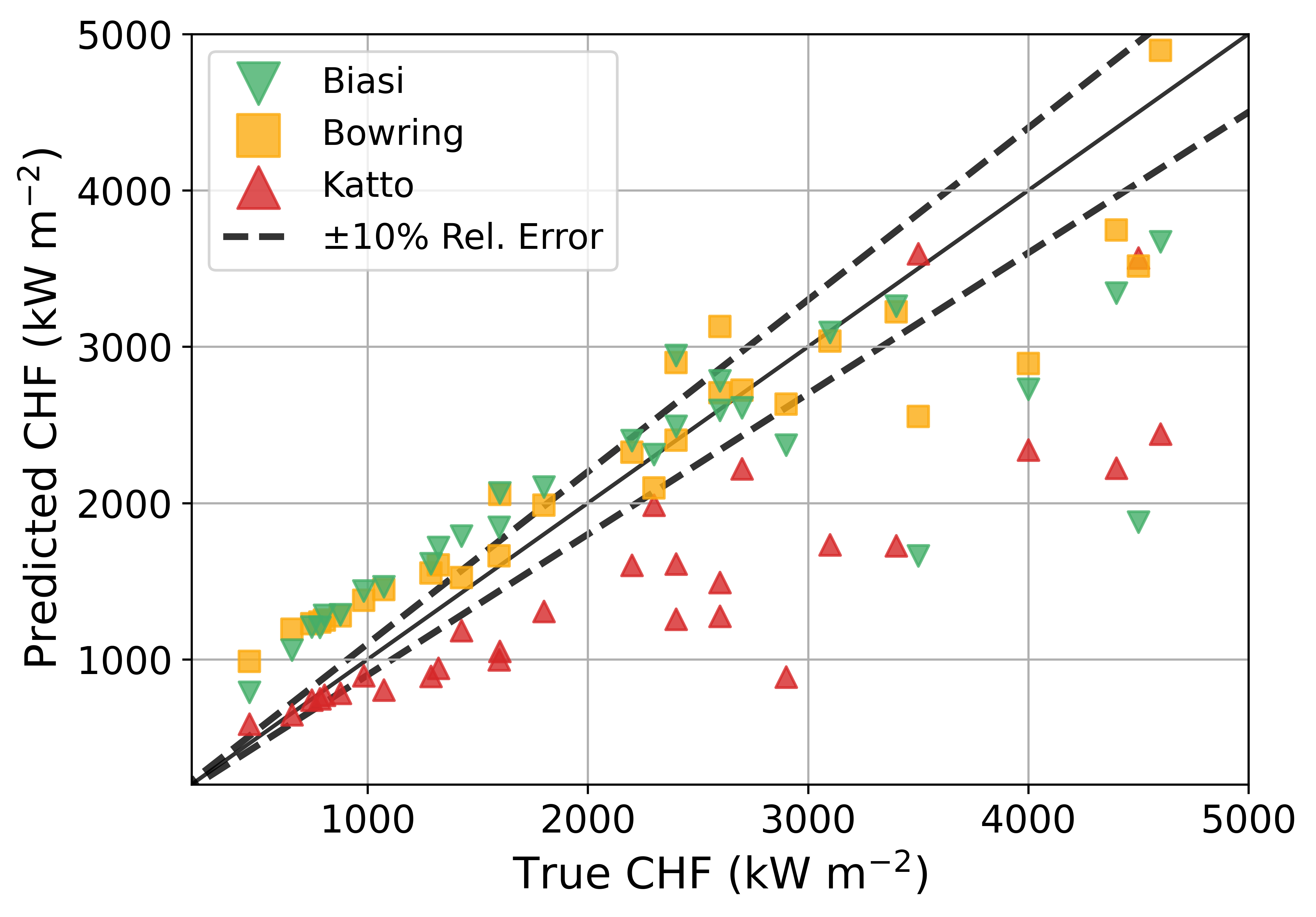}
    \caption{Parity of the standalone base models.}
    \label{fig:parity_base}
\end{figure}

All three empirical correlations produced mean relative errors between 26\% and 28\%, and more than 60\% of the test points exceeded a 10\% error threshold. The Bowring model exhibited the highest maximum relative error (113.34\%) compared to Biasi (70.66\%) and Katto (69.37\%). Although the performance of these correlations is quantitatively similar, significantly different error distributions were observed, as shown in Figure \ref{fig:parity_base}. All three models demonstrate tight clustering at smaller CHF values, with residuals increasing considerably at larger CHF values. While Katto is observed with favorable parity at smaller CHF values, it begins to significantly underpredict above 1,000 \si{\kilo\watt\per\square\meter}.

\begin{table}[htb!]
    \centering
    \caption{Performance of the standalone correlations on the annulus test partition.}
    \label{tab:comparison_base}
    \begin{tabular}{lccc}
        \toprule
        Metric & Base & Base & Base \\
          & Biasi & Bowring & Katto \\
        \midrule
        $\mu_\text{error}$ (\%)          & $28.83$ & $26.28$  & $28.06$ \\ \midrule
        $\text{Max}_{\text{error}}$ (\%) & $70.66$ & $113.34$ & $69.37$ \\ \midrule
        $\text{Std}_{\text{error}}$ (\%) & $21.87$ & $26.97$  & $18.03$ \\ \midrule
        $rRMSE$ (\%)                     & $35.96$ & $37.33$  & $33.18$ \\ \midrule
        $F_{\text{error}}>10\%$ (\%)     & $72.41$ & $62.07$  & $79.31$ \\
        \bottomrule
    \end{tabular}
\end{table}

CTF models with the same geometry as those used previously were then created for the ML-based cases. The same CHF extraction procedure was applied, and performance metrics were computed and are presented in Table \ref{tab:comparison_ml}. Compared to the base models, the ML models exhibit substantial improvement across all error metrics. All mean relative errors are below 3.5\%, and maximum relative errors do not exceed 13.1\%. The fraction of points exceeding 10\% is minimal; the worst cases correspond to a single test point (3.45\%).

\begin{figure}
    \centering
    \includegraphics[width=\linewidth]{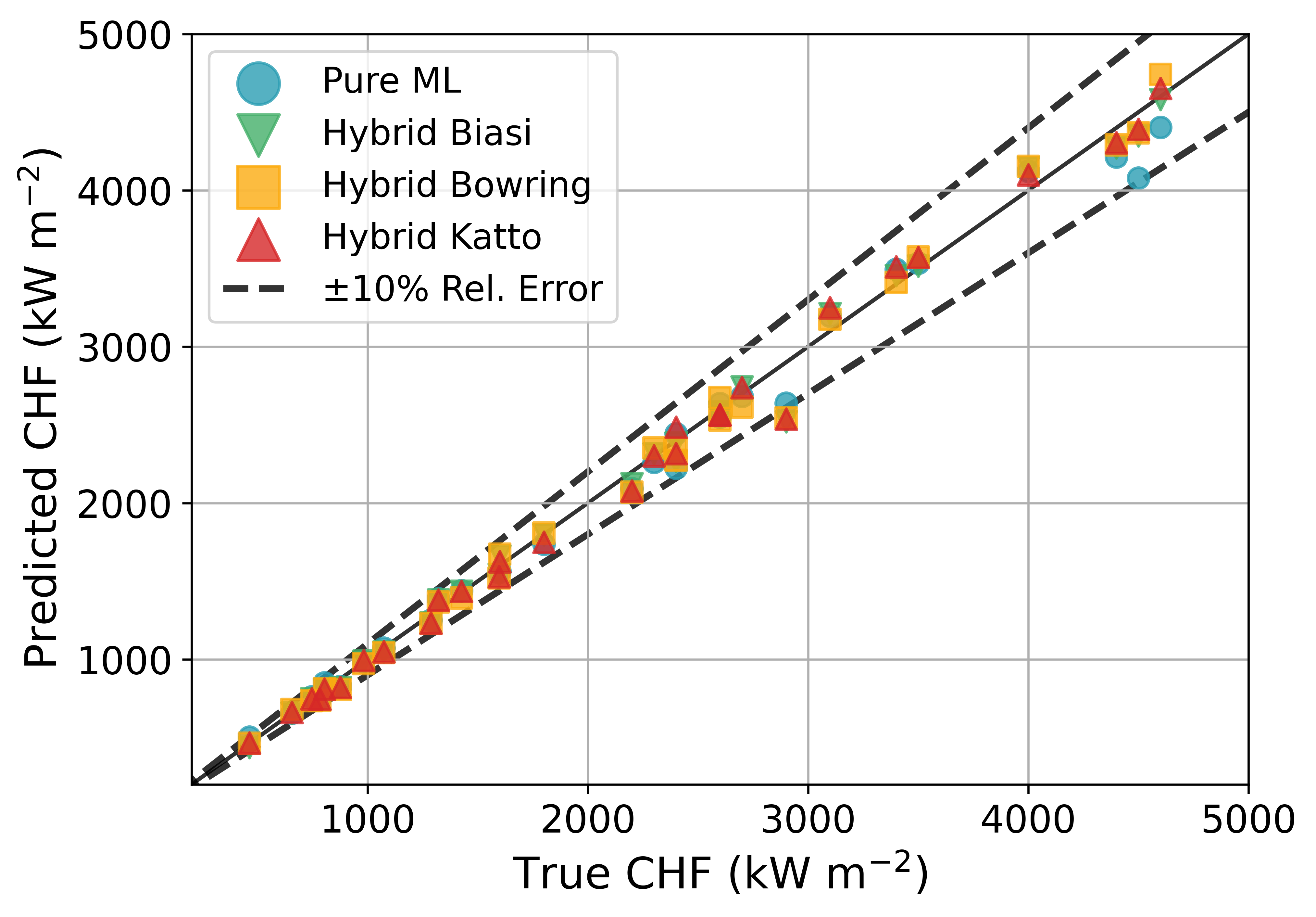}
    \caption{Parity of the pure and hybrid ML models.}
    \label{fig:parity_ml}
\end{figure}

No single ML model consistently outperforms the others across all metrics, making it difficult to determine a clear best performer. With respect to parity in Figure \ref{fig:parity_ml}, test points are symmetrically distributed about the identity line, indicating a lack of systematic bias. Only three points lie outside of the $\pm$10\% error envelope, one from each of the hybrid Biasi, hybrid Bowring, and hybrid Katto models. Additionally, five points in total exceed an absolute error of 200 \si{\kilo\watt\per\square\meter}.

\begin{table}[htb!]
    \centering
    \caption{Performance of the pure and hybrid ML variants on the annulus test partition.}
    \label{tab:comparison_ml}
    \begin{tabular}{lcccc}
        \toprule
        Metric & Pure & Hybrid & Hybrid & Hybrid \\
          & ML & Biasi & Bowring & Katto \\
        \midrule
        $\mu_\text{error}$ (\%)          & $3.47$ & \cellcolor{gray!25} $2.78$ & $3.33$   & $2.83$ \\ \midrule
        $\text{Max}_{\text{error}}$ (\%) & \cellcolor{gray!25} $9.36$ & $13.07$ & $12.19$ & $12.49$ \\ \midrule
        $\text{Std}_{\text{error}}$ (\%) & $2.62$ & $2.70$ & \cellcolor{gray!25} $2.46$   & $2.48$ \\ \midrule
        $rRMSE$ (\%)                     & $4.32$ & $3.84$ & $4.12$ & \cellcolor{gray!25} $3.74$ \\ \midrule
        $F_{\text{error}}>10\%$ (\%)     & \cellcolor{gray!25} $0.00$ & $3.45$ & $3.45$   & $3.45$ \\
        \bottomrule
    \end{tabular}
\end{table}

\section{Conclusions}

This study developed, deployed, and validated four ML models to predict CHF in annular geometries using the CTF subchannel code. Three models followed a hybrid residual-learning approach, in which physics-based empirical correlations were corrected using ML-predicted residuals between correlation estimates and experimental data. Three empirical correlation models, Biasi, Bowring and Katto, were used as base models for comparison.

The ML models were trained and tested using 577 experimental annulus data points from four datasets: Becker, Beus, Janssen, and Mortimore. A training-heavy split was used, reserving 29 points for testing. 

Baseline CHF predictions were obtained from the empirical correlations, with mean relative errors above 26\%. While quantitatively similar, the baseline models exhibited different error patterns and inconsistent performance across the test domain. The ML-driven models achieved mean relative errors below 3.5\%, with no more than one point exceeding the 10\% error envelope. Error distributions for all ML models were symmetric about the parity line, indicating no strong bias.

In all cases, the ML models significantly outperformed their empirical counterparts. More accurate CHF prediction improves dryout localization and enables the definition of more reliable thermal safety margins, which, in turn, support safer and more optimized system design. Although these annulus-specific models are primarily intended for application in annular geometries, they may be selected from a broader model library at the user’s discretion. Future work will focus on extending these models to rod bundle geometries in support of ongoing CTF development. Once that work is complete, model performance will be benchmarked against high-quality test cases, including the BFBT and CE $5\times5$ experiments.

\section{Acknowledgments}

\noindent 
This research was supported in part by an appointment to the U.S. Department of Energy's Omni Technology Alliance Internship Program, sponsored by DOE and administered by the Oak Ridge Institute for Science and Education. The authors from the North Carolina State University were also funded by the U.S. DOE Office of Nuclear Energy Distinguished Early Career Program (DECP) under award number DE-NE0009467.

\bibliographystyle{ans}
\bibliography{bibliography}

\begin{thebibliography}{10}
\newcommand{\enquote}[1]{``#1''}

\bibitem{zhao2020prediction}
\MakeUppercase{X.~Zhao}, \MakeUppercase{K.~Shirvan}, \MakeUppercase{R.~K. Salko}, and \MakeUppercase{F.~Guo}, \enquote{On the prediction of critical heat flux using a physics-informed machine learning-aided framework,} \emph{Applied Thermal Engineering}, \textbf{164}, 114540 (2020).

\bibitem{furlong2025deployment}
\MakeUppercase{A.~Furlong}, \MakeUppercase{X.~Zhao}, \MakeUppercase{R.~Salko}, and \MakeUppercase{X.~Wu}, \enquote{Deployment of Traditional and Hybrid Machine Learning for Critical Heat Flux Prediction in the CTF Thermal Hydraulics Code,} \emph{arXiv preprint arXiv:2505.14701} (2025).

\bibitem{salko2020ctf}
\MakeUppercase{R.~K. Salko~Jr}, \MakeUppercase{M.~Avramova}, \MakeUppercase{A.~J. Wysocki}, \MakeUppercase{J.~Hu}, \MakeUppercase{A.~Toptan}, \MakeUppercase{N.~Porter}, \MakeUppercase{T.~S. Blyth}, \MakeUppercase{C.~A. Dances}, \MakeUppercase{A.~Gomez}, \MakeUppercase{C.~Jernigan}, \MakeUppercase{et~al.}, \enquote{CTF User's Manual (V. 4.2),} Tech. rep., Oak Ridge National Lab.(ORNL), Oak Ridge, TN (United States); US Nuclear~… (2020).

\bibitem{biasi1967studies}
\MakeUppercase{L.~Biasi}, \enquote{Studies on burnout; Part 3-a new correlation for round ducts and uniform heating and its comparison with world data,} \emph{Energ. Nucl.(Rome)}, \textbf{14}, 530 (1967).

\bibitem{bowring1972simple}
\MakeUppercase{R.~Bowring}, \enquote{A simple but accurate round tube, uniform heat flux, dryout correlation over the pressure range 0.7-17 MN/m 2 (100-2500 PSIA),} Tech. rep., UKAEA Reactor Group (1972).

\bibitem{katto1979generalized}
\MakeUppercase{Y.~Katto}, \enquote{Generalized correlations of critical heat flux for the forced convection boiling in vertical uniformly heated annuli,} \emph{International Journal of Heat and Mass Transfer}, \textbf{22}, \emph{4}, 575--584 (1979).

\bibitem{hejzlar1996consideration}
\MakeUppercase{P.~Hejzlar} and \MakeUppercase{N.~E. Todreas}, \enquote{Consideration of critical heat flux margin prediction by subcooled or low quality critical heat flux correlations,} \emph{Nuclear engineering and design}, \textbf{163}, \emph{1-2}, 215--223 (1996).

\bibitem{furlong2025native}
\MakeUppercase{A.~Furlong}, \MakeUppercase{X.~Zhao}, \MakeUppercase{B.~Salko}, and \MakeUppercase{X.~Wu}, \enquote{Native Fortran Implementation of TensorFlow-Trained Deep and Bayesian Neural Networks,} \emph{arXiv preprint arXiv:2502.06853} (2025).

\bibitem{becker1981experimental}
\MakeUppercase{K.~M. Becker}, \MakeUppercase{A.~Letzter}, \MakeUppercase{O.~Nylund}, and \MakeUppercase{B.~Sch{\"o}lin}, \enquote{An experimental study of the effect of the axial heat flux distribution on the dryout conditions in a 3650-mm long annulus,} \emph{International Journal of Multiphase Flow}, \textbf{7}, \emph{1}, 47--61 (1981).

\bibitem{beus1981critical}
\MakeUppercase{S.~Beus} and \MakeUppercase{O.~Seebold}, \enquote{Critical heat flux experiments in an internally heated annulus with a non-uniform, alternate high and low axial heat flux distribution (AWBA Development Program),} Tech. rep., Bettis Atomic Power Lab.(BAPL), West Mifflin, PA (United States) (1981).

\bibitem{janssen1963burnout}
\MakeUppercase{E.~Janssen} and \MakeUppercase{J.~Kervinen}, \enquote{Burnout conditions for single rod in annular geometry, water at 600 to 1400 psia,} Tech. rep., General Electric Co. Atomic Power Equipment Dept., San Jose, Calif. (1963).

\bibitem{mortimore1979critical}
\MakeUppercase{E.~Mortimore} and \MakeUppercase{S.~G. Beus}, \enquote{Critical heat flux experiments with a local hot patch in an internally heated annulus (LWBR development program),} Tech. rep., Bettis Atomic Power Lab.(BAPL), West Mifflin, PA (United States) (1979).

\end{thebibliography}
\end{document}